\pdfoutput=1

\documentclass[11pt]{article}

\usepackage[preprint]{acl}

\usepackage{times}
\usepackage{latexsym}

\usepackage[T1]{fontenc}

\usepackage[utf8]{inputenc}

\usepackage{microtype}

\usepackage{inconsolata}

\usepackage{graphicx}
\usepackage{amsmath}
\usepackage{booktabs}
\usepackage{multirow}

\newcommand{\newshortname}{VisualRWKV}

\title{VisualRWKV: Exploring Recurrent Neural Networks for Visual Language Models}

\author{Haowen Hou\textsuperscript{*}
  \and 
  Peigen Zeng\textsuperscript{+}
  \and 
  Fei Ma\textsuperscript{*}
  \and 
  Fei Richard Yu\textsuperscript{+\textdagger} \\
  \textsuperscript{*}Guangdong Laboratory of Artificial Intelligence and Digital Economy (SZ), Shenzhen, China \\
  \textsuperscript{+}College of Computer Science and Software Engineering, Shenzhen University, Shenzhen, China \\
  \textsuperscript{\textdagger}Shool of Information Technology, Carleton University, Canada \\
  \texttt{\{houhaowen, mafei, yufei\}@gml.ac.cn}
  \thanks{This work is supported in part by Shenzhen Science and Technology Program under Grant ZDSYS20220527171400002, the National Natural Science Foundation of China (NSFC) under Grants 62406197, 62271324, 62231020 and 62371309. Corresponding author: F. Richard Yu.
	}
  }

\begin{document}
\maketitle

\begin{abstract}
Visual Language Models (VLMs) have rapidly progressed with the recent success of large language models. However, there have been few attempts to incorporate efficient linear Recurrent Neural Networks (RNNs) architectures into VLMs.
In this study, we introduce VisualRWKV, the first application of a linear RNN model to multimodal learning tasks, leveraging the pre-trained RWKV language model. 
We propose a data-dependent recurrence and sandwich prompts to enhance our modeling capabilities, along with a 2D image scanning mechanism to enrich the processing of visual sequences.
Extensive experiments demonstrate that VisualRWKV achieves competitive performance compared to Transformer-based models like LLaVA-1.5 on various benchmarks.
Compared to LLaVA-1.5, VisualRWKV has a speed advantage of 3.98 times and can save 54\% of GPU memory when reaching an inference length of 24K tokens.
To facilitate further research and analysis, we have made the checkpoints and the associated code publicly accessible at the following GitHub repository: \href{https://github.com/howard-hou/VisualRWKV}{https://github.com/howard-hou/VisualRWKV}.
\end{abstract}

\section{Introduction}
Large Language Models (LLMs) have demonstrated exceptional performance in natural language processing tasks \cite{touvron2023llama2,brown2020language}. 
Extending LLMs to support visual inputs has garnered significant attention in the research community \cite{gpt4v}. 
Visual Language Models (VLMs) inherit powerful capabilities from LLMs, such as strong instruction following, zero-shot generalization, and in-context learning \cite{liu2023llava,zhu2023minigpt}. 
By integrating visual and textual information, VLMs not only enhance the understanding of visual content but also provide richer context for language understanding and generation. 
VLMs hold tremendous potential for solving visual problems and advancing various vision-language tasks. 

\begin{figure}[t!]
    \centering
    \includegraphics[width=0.9\linewidth]{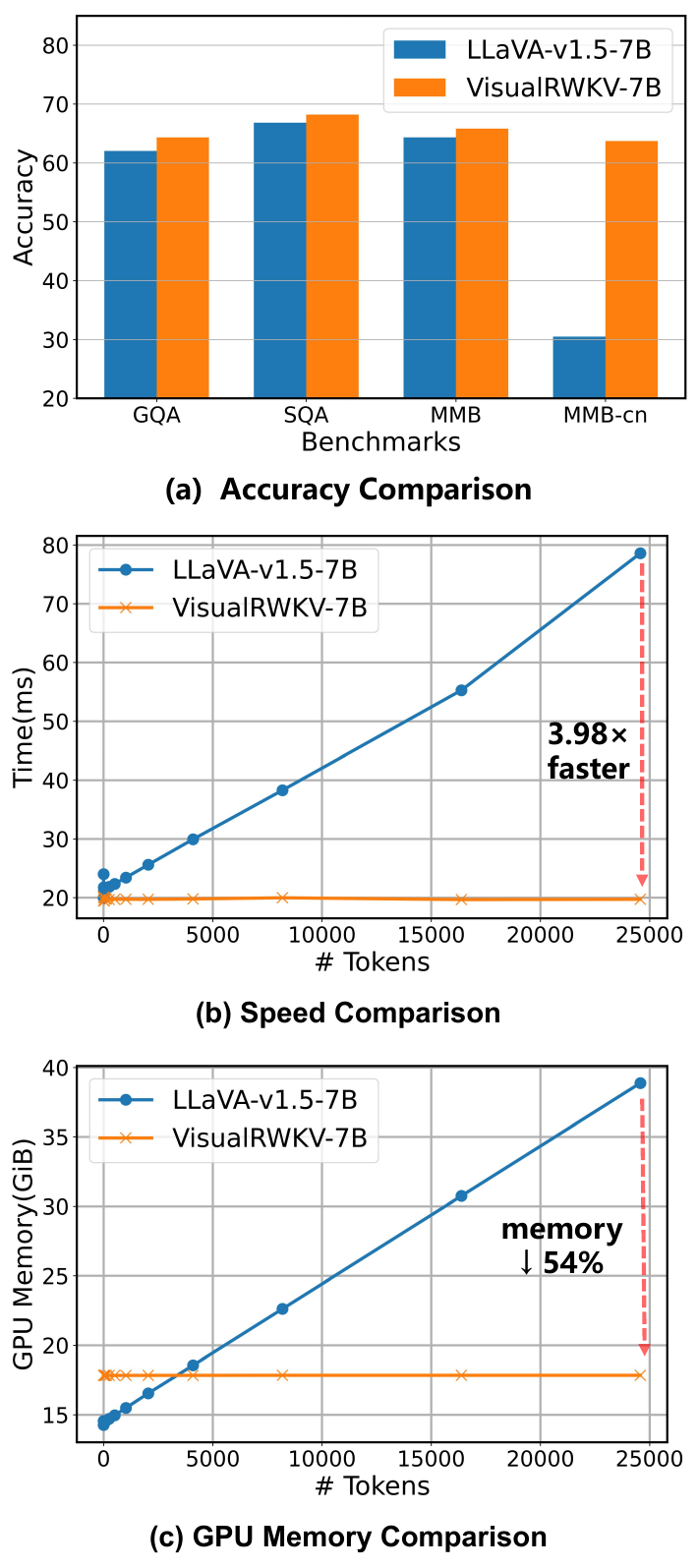}
    \caption{\textbf{VisualRWKV} outperforms the SoTA LLaVA-1.5 \citep{liu2023improvedllava} on 4 tasks (a), with high computational efficiency (b) and low, stable memory usage (c).}
    \vspace{-.2 in}
    \label{fig:teaser}
\end{figure}

However, despite the excellent performance of existing LLMs and VLMs, their inherent computational and memory complexity due to the self-attention mechanism in the Transformer architecture results in quadratic growth in computation and memory requirements with the increase in sequence length \cite{katharopoulos2020lineartransformrers}. This leads to high inference costs and limits the deployment and application of Transformer-based VLMs on edge devices.

The Receptance Weighted Key Value (RWKV) model, a novel Recurrent Neural Network (RNN) architecture, presents a promising solution to the bottleneck of long-sequence modeling \cite{peng2023rwkv}. It surpasses Transformers in large-scale data performance and exhibits linear scalability with sequence length, positioning itself as a promising successor to Transformers in language modeling \cite{Peng2023RWKVRR}.

Currently, there is a notable gap in research exploring how this efficient architecture can be leveraged for multimodal tasks. In this study, we introduce the VisualRWKV model, marking the first application of a linear RNN model to multimodal learning tasks. Specifically, we utilize the pre-trained RWKV language model as the foundational language model and explore several novel mechanisms applied to VisualRWKV.

VisualRWKV introduces: 
(1) an innovative data-dependent recurrence to enhance the capabilities and capacity of the RWKV model. 
(2) a novel sandwich prompt designed to provide richer conditions when processing visual sequences.
(3) a new 2D image scanning mechanism to enhance the 2D modeling capabilities of visual sequences.

Extensive experiments on various multimodal learning benchmarks validate the effectiveness of VisualRWKV, as shown in Figure \ref{fig:teaser}. Compared to other Transformer-based models of similar size, such as LLaVA-1.5 \cite{liu2023improvedllava}, VisualRWKV demonstrates competitive performance, achieving outstanding results on multiple popular benchmarks.

In summary, this study presents the VisualRWKV model, explores the impact of various novel designs on VisualRWKV, introduces the innovative sandwich prompt to enhance representation capabilities, and conducts extensive experiments across diverse multimodal learning benchmarks.

\section{Related Works}
\subsection{Visual Language Models}
Following the success of LLMs, recent research has pivoted towards VLMs \citep{achiam2023gpt, team2023gemini} for enhancing visual understanding and reasoning capabilities. 
Expanding on various pre-trained LLM architectures, researchers have proposed diverse methodologies for incorporating visual information. Flamingo \citep{alayrac2022flamingo} and BLIP-2 \citep{li2023blip} introduce distinct techniques for modality fusion, integrating visual tokens with frozen large language models through gated attention or query transformers. 
Building on the effectiveness of instruction tuning, LLaVA \citep{liu2023llava, liu2023improvedllava} and MiniGPT-4 \citep{zhu2023minigpt, chen2023minigptv2} utilize visual instruction tuning to align visual input with LLMs, showcasing notable achievements. 
Recent advancements, such as Kosmos-2 \citep{peng2023kosmos} and Shikra \citep{chen2023shikra}, further enhance VLMs with grounded visual understanding capabilities. 
Despite their promising potential for general-purpose visual reasoning and planning tasks, these models are generally expensive and challenging to train and deploy.

\subsection{Linear RNN Large Language Model}
Recent advancements in LLMs, such as GPT \citep{radford2019language, brown2020language, achiam2023gpt}, LLaMA \citep{touvron2023llama,touvron2023llama2}, and PaLM \citep{anil2023palm, chowdhery2023palm}, have showcased remarkable prowess across various natural language processing tasks. However, traditional Transformer-based LLMs suffer from quadratic complexity $O(L^2)$ issues in both computation and memory, prompting the emergence of linear RNNs as potential successors.

RNNs model sequential data with temporal dependencies by generating a hidden state $h_t$ at each time step, which is then utilized as input for the subsequent step. Classical RNN variants like LSTM~\citep{hochreiter1997long} and GRU~\citep{cho2014learning} excel in inexpensive inference, operating typically at $O(1)$ time complexity per step relative to sequence length. Nonetheless, their older designs often pose challenges in parallelization across time dimensions during training.

Linear RNNs present themselves as promising successors to the Transformer, offering a more efficient token mixing method. They enable a space complexity of $O(L)$ and an inference complexity of $O(1)$. Leveraging Parallel Prefix Sum Scan \citep{Harris2011ParallelPS} for acceleration can further enhance their efficiency. The RWKV \citep{Peng2023RWKVRR,Hou2024RWKVTSBT}, a linear RNN-based LLM, has showcased competitive performance compared to GPT models of similar scale. RWKV introduces temporal decay to gradually reduce the influence of past information, implicitly incorporating positional information. Additionally, it integrates a token-shift mechanism facilitating linear interpolation between current and previous inputs. This allows the model to naturally aggregate and regulate information within input channels. Furthermore, RWKV boasts a time complexity of $O(L)$ and an inference complexity of $O(1)$, ensuring consistent inference time per token. As a result, the overall inference duration scales linearly with sequence length. The memory footprint of RWKV remains constant, regardless of sequence length, contributing to its efficiency and scalability.

\section{Methods}
In this section, we initially introduce the fundamental concepts of the RWKV language model. (Section \ref{sec:preliminaries}).
Following that, we elaborate on the transformation of the RWKV language model into our proposed VisualRWKV visual language model (Section \ref{sec:visualrwkv}), which mainly includes data-dependent recurrence, sandwich prompting, and image scanning. 
\begin{figure*}[hbt]
    \centering
    \includegraphics[width=0.95\linewidth]{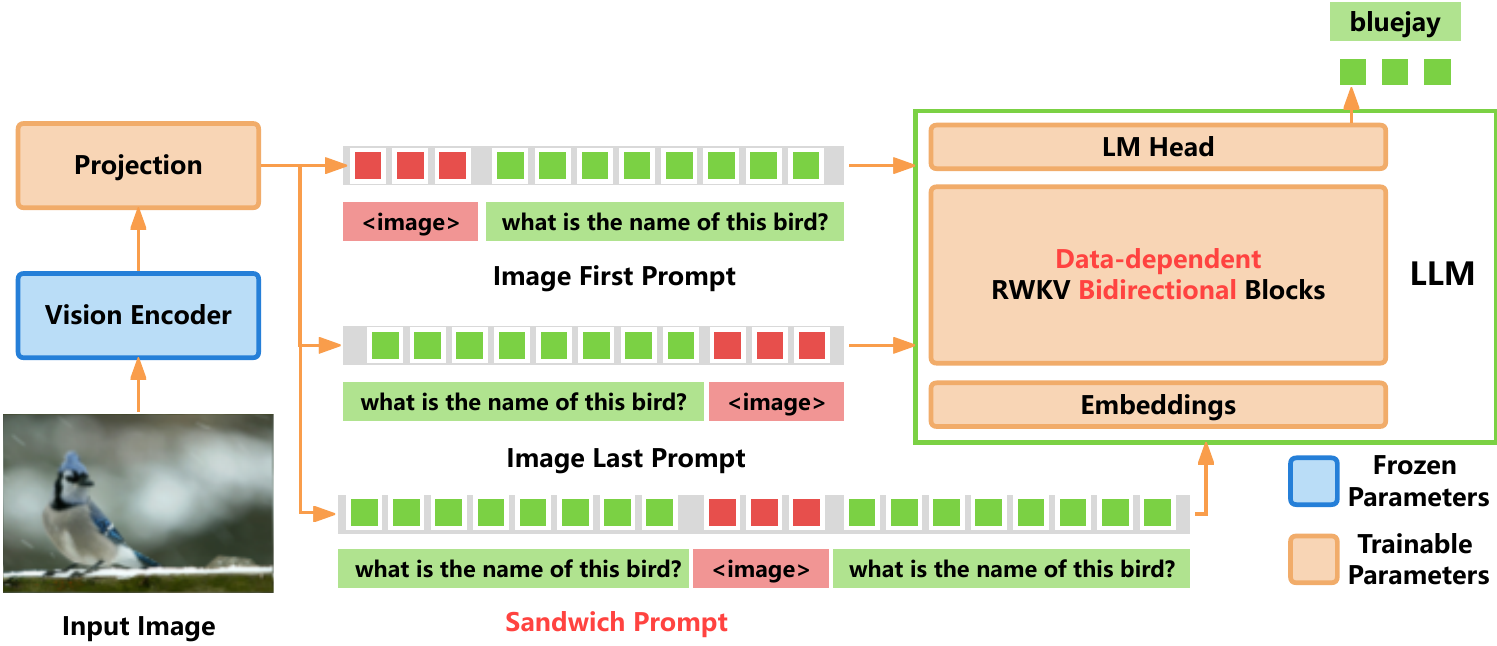}
    \caption{VisualRWKV architecture overview and three prompting method. \textbf{Image First Prompt:} place image tokens before instruction tokens; \textbf{Image Last Prompt:} place image tokens after instruction tokens; \textbf{Sandwich Prompt:} place image tokens in the middle of instruction tokens. Red words indicate the key contributions.}
    \label{fig:visualrwkv_arch}
\end{figure*}

\subsection{Preliminaries}
\label{sec:preliminaries}

The RWKV\citep{Peng2024EagleAF} backbone is structured using stacked residual blocks, with each block containing a time-mixing and a channel-mixing sub-block. These components embody recurrent structures designed to leverage past information.

\paragraph{Data-independent Token Shift}

As shown in Figure \ref{fig:dynamic}, trainable variable $\mu_g$, $\mu_r$, $\mu_k$, $\mu_v$ are used in a linear combination of $x_t$ and $x_{t-1}$, to achieve a simple mixing, which interpolate between the inputs of the current and previous time-steps. The combination of shifted previous step and current step was linear projected through projection matrix within the block:
\begin{equation} 
\label{eq:token-shift}
\small
\alpha_{t} = (\mu_{\alpha} \odot x_t + (1 - \mu_{\alpha}) \odot x_{t-1} )W_{\alpha}  \\
\end{equation}
where $\alpha$ serves as a notation for the variables $r$, $g$, $k$, and $v$, given that they are subject to an identical linear combination formula. Please note that the linear combination used here is data independent, meaning the value of $\mu_\alpha$ is not dependent on $x_t$ or $x_{t-1}$.

\paragraph{Data-independent Time Mixing}

In vanilla RWKV, the time mixing is articulated through the update of the WKV vectors and the WKV operator is input-data independent. The formula of single head WKV operator is given by:
\begin{equation}
    \label{eq:data-indep-wkv}
    \small
    wkv_{t} = \mathrm{diag}(u)\cdot k_{t}^\mathrm{T} \cdot v_{t} + \sum_{i=1}^{t-1} \mathrm{diag}(w)^{t-1-i} \cdot  k_{i}^\mathrm{T} \cdot v_{i} \\
\end{equation}
where $w$ and $u$ are two trainable parameters. The parameter $u$ serves as a term weight for the current token when the model encounters it for the first time. It enables the model to efficiently process the token by focusing more on important tokens and quickly filtering out unimportant ones. Another important parameter is $w$, which is a channel-wise time decay vector per head. Furthermore, we transform parameter $w$ by $w = \exp(-\exp(w)) $.
This transformation ensures that all values of ${w}$ are within the range $(0,1)$, ensuring that $\mathrm{diag}(w)$ represents a contraction matrix.

The output from the single-head WKV operator undergoes processing by the layer normalization and the SiLU activation. Then, all outputs are concatenated to form the output vector $o_t$:
\begin{equation}
    \small
    o_t = concat( SiLU(g_t) \odot \mathrm{LayerNorm}(r_t \cdot wkv_t) )W_o
\end{equation}
where LayerNorm operates on each head separately. For further details and formulas of the models, one can refer to \citet{Peng2024EagleAF} and \citet{Hou2024RWKVTSBT}.

\subsection{VisualRWKV}
\label{sec:visualrwkv}

\begin{table}[h!]
\centering
\resizebox{\columnwidth}{!}{%
\begin{tabular}{@{}llllll@{}}
\toprule
Method                 & Size & VQA   & SQA   & TQA   & GQA   \\ \midrule
VisualRWKV-Base    & 1.6B & 51.08 & 41.94 & 35.19 & 48.09 \\
+Data-dep Recurrence   & 1.6B & 65.82 & 46.55 & 40.26 & 49.06 \\
+Bidirection +Sandwich & 1.6B & 64.96 & 56.72 & 41.94 & 48.04 \\
+Better Learning Rate  & 1.6B & 69.42 & 59.05 & 43.57 & 55.23 \\
+Scale up to 3B   & 3B   & 71.52 & 65.34 & 48.68 & 59.56 \\ 
+Scale up to 7B   & 7B   & 75.82 & 68.22 & 51.01 & 64.27 \\ \bottomrule
\end{tabular}%
}
\caption{
\textbf{Scaling results} on model. We choose to conduct experiments on VQA-v2(VQA), ScienceQA(SQA), TextVQA(TQA) and GQA to examine model's capabilities.}
\label{tab:scaling_ablation}
\end{table}

\subsubsection{VisualRWKV Baseline}
\begin{figure}[ht]
    \centering
    \includegraphics[width=\linewidth]{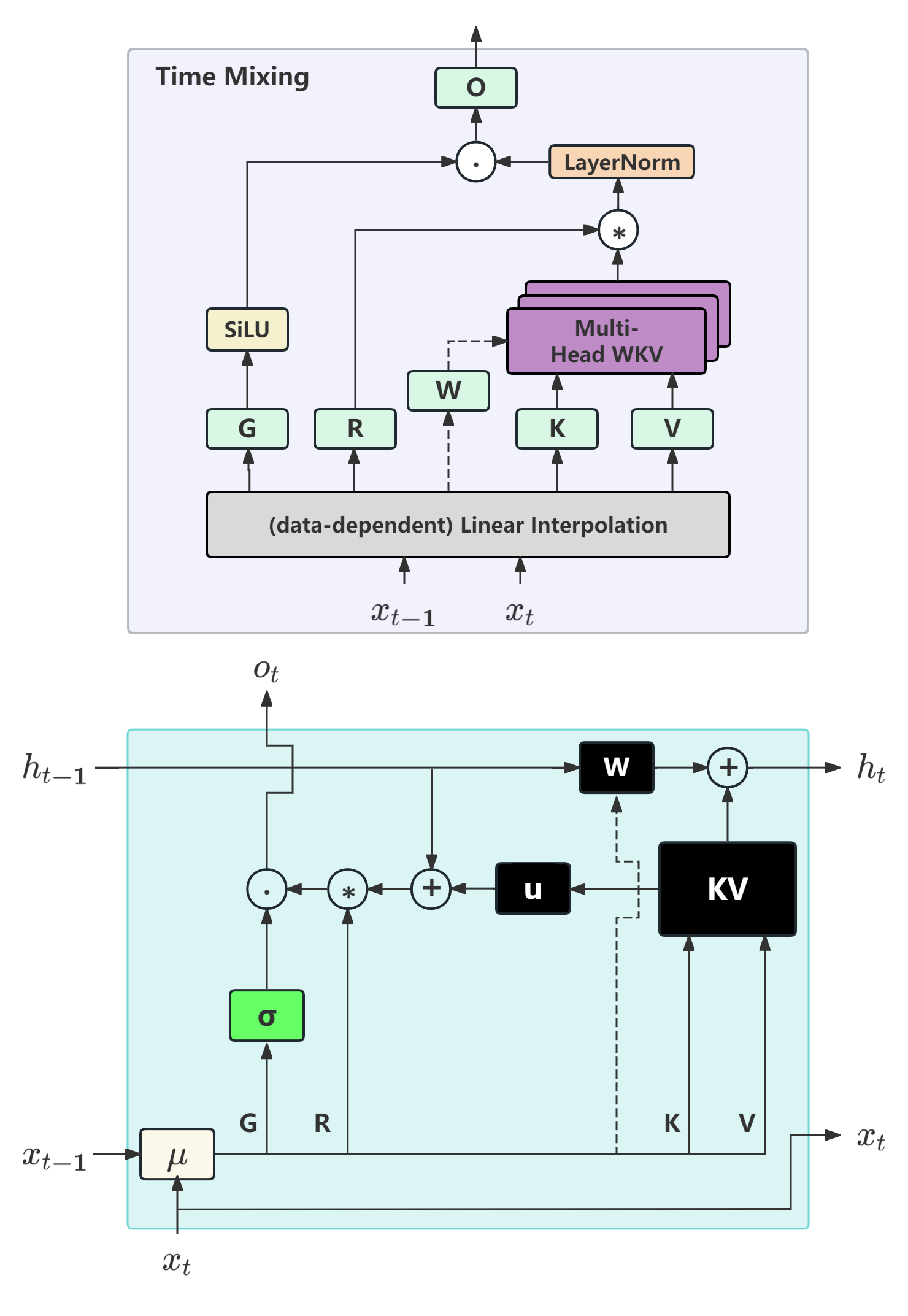}
    \caption{Data-dependent recurrence. \textbf{Top:} Semantic diagram of the time-mixing block; \textbf{Bottom:} Time-mixing block as an RNN cell. Dashed arrows represent connections in data-dependent recurrence, not present in data-independent recurrence.}
    \label{fig:dynamic}
\end{figure}

VisualRWKV is a follow-up work to RWKV.
RWKV paper \citep{Peng2024EagleAF} proposed a simplified version of VisualRWKV that employed data-independent recurrence (Fig. \ref{fig:dynamic}), unidirection image scanning (Fig. \ref{fig:image_scanning}), and image first prompting (Fig. \ref{fig:visualrwkv_arch}).
We used that version of VisualRWKV as the baseline and starting point for our research, as shown in Table \ref{tab:scaling_ablation}. We denote this initial model without any modifications as \textbf{VisualRWKV-Base}.

\subsubsection{Data-dependent Recurrence}
\label{sec:data-dep}
The Data-dependent Recurrence mechanism introduces two key enhancements: the Data-dependent Token Shift and the Data-dependent Time Mixing.

\paragraph{Data-dependent Token Shift}
First, we define \textbf{lo}w-\textbf{r}ank \textbf{a}daptation (lora) and \textbf{d}ata-\textbf{d}ependent \textbf{l}inear int\textbf{erp}olation (ddlerp) as follow:
\begin{equation}
    \small
    \mathrm{lora}_\alpha(x)=\lambda_\alpha+\tanh(xA_\alpha)B_\alpha
\end{equation}
\begin{equation}
    \small
    \mathrm{ddlerp}_\alpha(a, b) = a + (b-a) \odot \mathrm{lora}_\alpha(a + (b-a) \odot \mu_x)
\end{equation}

Then, the Data-dependent Token Shift is defined as:
\begin{equation}
    \small
    \alpha_{t} = \mathrm{ddlerp}_{\alpha}(x_t, x_{t-1})W_{\alpha}
\end{equation}
where $\alpha$ serves as a notation for the variables $r$, $g$, $k$, and $v$. $A_\alpha$, $B_\alpha$, $\lambda_\alpha$ and $W_\alpha$ are trainable parameters. The data-dependent token shift seeks to broaden the model's capacity. It dynamically allocates the ratio of new to existing data per channel, depends on the input at both current and previous time steps.

\paragraph{Data-dependent Time Mixing}

The key improvement over data-independent time mixing (Eq. \ref{eq:data-indep-wkv}) lies in the evolution of the time decay vector from a fixed parameter $w$ to a dynamic one $w_t$ that reacts to the input data $x_t$ at time step $t$.
The dynamic nature of $w_t$ allows the model to adjust more nimbly to diverse input data, unbound by rigid, predefined structures. Equations are as follow:

\begin{equation}
    \small
    d_t = \mathrm{lora}_d( \mathrm{ddlerp}_d ( x_t, x_{t-1} ) )
\end{equation}
\begin{equation}
    \small
     w_t = \exp(-\exp(d_t))
\end{equation}
\begin{equation}
    \label{eq:data-dep-wkv}
    \small
    wkv_{t} = \mathrm{diag}(u)\cdot k_{t}^\mathrm{T} \cdot v_{t} + \sum_{i=1}^{t-1}  \mathrm{diag}\left(\bigodot_{j=1}^{i-1}w_{j}\right) \cdot  k_{i}^\mathrm{T} \cdot v_{i}
\end{equation}

The LoRA mechanism utilizes vectors learned from data-independent time mixing and enhances them at a low cost with additional offsets modulated by the incoming input.
It should be noted that the computation of the new time-varying decay $w_t$ employs a token-shifted value $\mathrm{ddlerp}_d ( x_t, x_{t-1} )$ as its input, not just the current token $x_t$.
As shown in Table \ref{tab:scaling_ablation}, the VisualRWKV equipped with data-dependent recurrence exhibits a significant improvement in performance.

\subsubsection{Sandwich Prompt}

The motivation for designing the sandwich prompt is as follows: Unlike the attention mechanism in Transformers, RNN models such as RWKV, due to their sequential nature, cannot revisit historical information repeatedly. 
Instead, they must decide immediately whether to store a token or image token in memory upon encountering it.
Therefore, carefully designing tailored prompts is essential for enhancing VisualRWKV's ability to effectively acquire and utilize information.
For this purpose, we have specifically designed three types of prompting methods, as shown in Figure \ref{fig:visualrwkv_arch}:
\begin{itemize}
    \item Image First Prompt: Place image tokens prior to the instruction tokens. 
    \item Image Last Prompt: Place image tokens following the instruction tokens.
    \item Sandwich Prompt: Insert image tokens between the instruction tokens.
\end{itemize}
The sandwich prompt is designed to provide optimal conditions that assist the model in making these decisions more effectively. 
Specifically, the first prompt helps the model efficiently extract relevant information from the image, while the second prompt focuses on improving the model’s ability to answer questions.

For instance, the Image Last Prompt can cause the model to occasionally forget the question embedded in the prompt, while the Image First Prompt may result in the model processing the image without considering the question, hindering its ability to analyze the image contextually. 
In contrast, the sandwich prompt resolves these issues and achieves a synergistic effect, enabling the model to perform better than the sum of the individual prompts.
The experimental results show that the Sandwich Prompt achieves the best performance, as presented in Table~\ref{tab:prompt_res}.

\subsubsection{Image Scanning}
\begin{figure*}[hbt]
    \centering
    \includegraphics[width=0.9\linewidth]{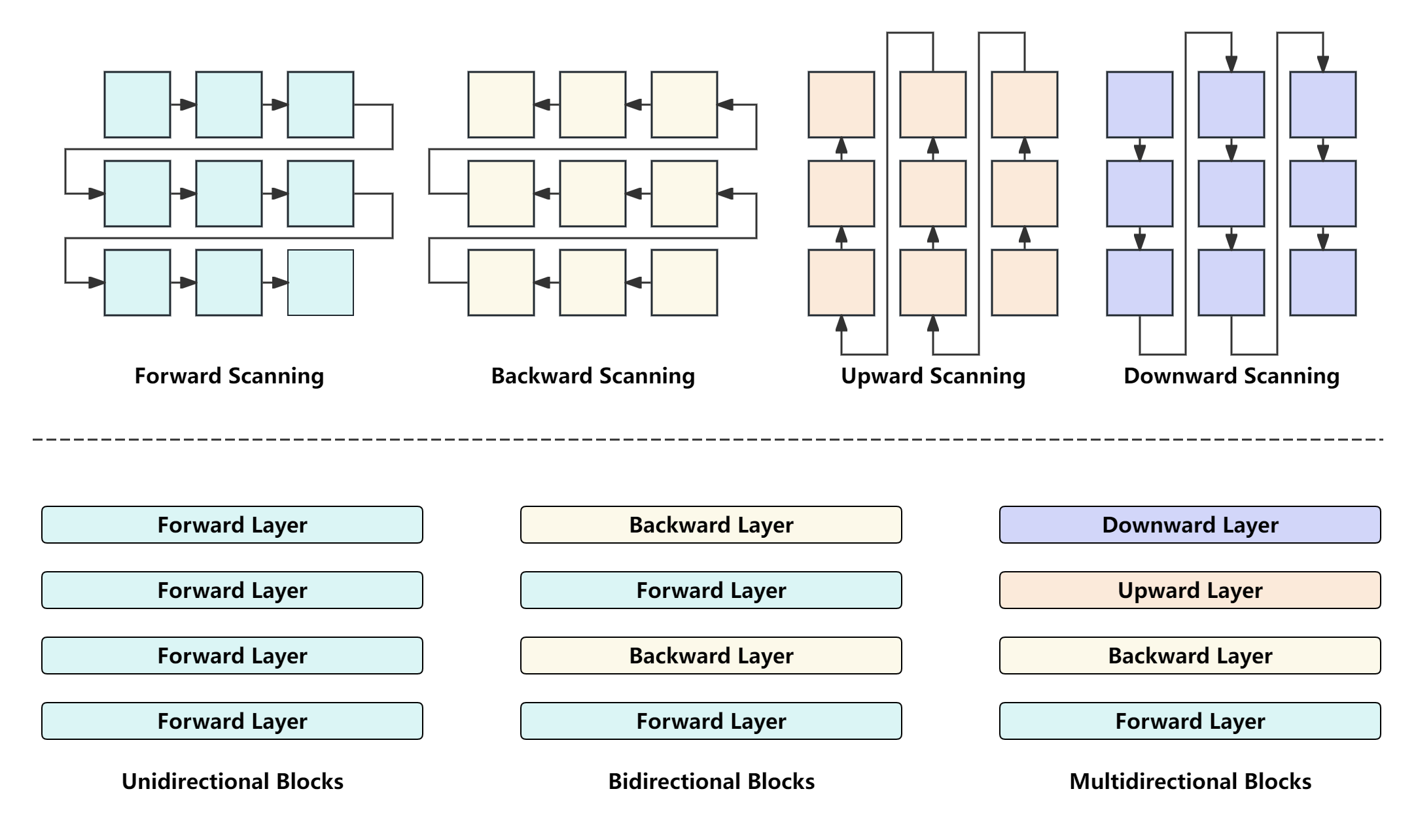}
    \caption{Illustration of 3 different multimodal RWKV Blocks: Unidirectional Blocks (left), Bidirectional Blocks (middle), and Multidirectional Blocks (right). The four scanning modes are also depicted at the top.}
    \label{fig:image_scanning}
\end{figure*}

The motivation for designing the image scanning techniques is as follows: Language is inherently unidirectional, while images are multidirectional by nature. As a result, unidirectional language models face inherent limitations when processing visual information. By implementing bidirectional or multidirectional image scanning strategies, these challenges can be effectively mitigated.

Vanilla RWKV is designed for 1D sequential data with causal relationships, such as language sequences. However, the visual sequences generated by vision encoders are non-causal. To bridge this gap, we propose a 2D scanning mechanism to improve VisualRWKV's performance on visual tasks. This work integrates the 2D scanning mechanism into RWKV blocks, exploring three variants of multimodal RWKV blocks, which are illustrated in Figure~\ref{fig:image_scanning}:

\begin{itemize}
    \item Unidirectional Blocks: Only containing the Forward Scanning Block, which is the basic scanning pattern of RWKV and other linear RNN models. This serves as the Base.
    \item Bidirectional Blocks: Comprising both Forward Scanning and Backward Scanning Blocks, arranged in an alternating fashion.
    \item Multidirectional Blocks: Including blocks for Forward Scanning, Backward Scanning, Upward Scanning, and Downward Scanning, with the sequence of Forward, Backward, Upward, and Downward arranged in an alternating order.
\end{itemize}

Our design alternates different scanning directions within layers, which does not introduce additional computational overhead and preserves the efficiency of the architecture.
The experimental results (Table~\ref{tab:scan_res}) have also verified the effectiveness and necessity of such scanning techniques in enhancing the model's ability to handle and understand visual sequences, thereby improving the overall performance of VisualRWKV in visual language processing tasks.

\section{Experiments}
The following section is dedicated to showcasing the key experiments and outcomes related to VisualRWKV. All results presented in this section are derived from a single run.

\subsection{Experiment Setup}
Following \citet{liu2023improvedllava,liu2023llava}, the training process of VisualRWKV consists of two stages: vision-and-language alignment pretraining and visual instruction tuning. In the pretraining stage, the vision encoder and RWKV LLM are frozen, with only the projector being updated. During the visual instruction tuning stage, we finetune both the projector and the RWKV LLM, as shown in Figure \ref{fig:visualrwkv_arch}. Details of training data and hyper-parameters can be found in Appendix \ref{sec:app_data_Hyperparameters}.

\subsection{Benchmarks}
We evaluated VisualRWKV across 8 benchmark tests tailored to assess the model's performance in academic tasks.

For assessing visual perception capabilities, VQA-v2 \citep{goyal2017vqav2} and GQA \citep{hudson2019gqa} presented open-ended short questions. Following the methodology outlined in LLaVA \citep{li2023llava}, we utilized the image subset of ScienceQA \citep{lu2022learn} to gauge the model's zero-shot generalization in answering scientific questions via multiple-choice questions. TextVQA \citep{singh2019textvqa} focused on visual question answering with rich text content.

Regarding benchmarks tailored for VLMs, various assessments evaluated the model's performance across diverse domains and applications, encompassing different response formats. MME-Perception \citep{fu2023mme} scrutinized the model's visual perception abilities through true/false questions. MMBench \citep{liu2023mmbench} assessed the robustness of the model's answers by rigorously shuffling multiple-choice options. MMBench-CN, the Chinese counterpart of MMBench, was employed to evaluate the model's multilingual capabilities. POPE \citep{li2023pope} assesses the model's hallucination degree on three sampled subsets of COCO \citep{lin2014mscoco}: random, common, and adversarial, reporting the average F1 score across all three splits.

\subsection{Quantitative Evaluation}
\subsubsection{Main Results}
Table \ref{tab:results} presents a comparison of our proposed VisualRWKV model with some state-of-the-art (SOTA) multimodal large language models.
VisualRWKV achieved the best performance in 3 out of 8 benchmarks and came in second place in SQA benchmark.
Compared with LLaVA-1.5 7B, which has similar scale parameters and the same amount of multimodal training data, Our model(VisualRWKV-7B) outperformed it in 4 benchmarks: SQA (68.2 vs. 66.8), GQA (64.3 vs. 62.0), MMB (65.8 vs. 64.3), and MMB-cn (63.7 vs. 30.5).
It is noteworthy that VisualRWKV and LLaVA-1.5 used completely identical training data.
Yet, on the MMB-cn Chinese test set, VisualRWKV showed a substantial lead.
This may indicate that the RWKV language model has stronger multilingual capabilities.
These promising results not only confirm the effectiveness of the VisualRWKV model, but also highlight the significant potential of the Linear RNN model in multimodal learning tasks.

\begin{table*}[ht]
\centering
\vspace{3pt}
\renewcommand{\arraystretch}{1.25}
\resizebox{\linewidth}{!}{
\begin{tabular}{ll cc | cccc | cccc }
\toprule
Method & LLM  & Res. & PT/IT & VQA & GQA & SQA & TQA & POPE & MME & MMB & MMB-cn \\
\midrule
BLIP-2~\citep{li2023blip} & Vicuna-13B  & 224 & 129M/ - & 41.0 & 41.0 & 61.0 & 42.5 & 85.3 & 1293.8 & -- & 22.4 \\
MiniGPT-4~\citep{zhu2023minigpt}&Vicuna-7B&224&5M/5K&-&32.2&-&-&-&581.7&23.0&-\\
InstructBLIP~\citep{Dai2023InstructBLIPTG} & Vicuna-7B & 224&129M/1.2M & -- & 49.2 & 60.5 & 50.1 & -- & -- & 36 & 26.2 \\ 
InstructBLIP~\citep{Dai2023InstructBLIPTG} & Vicuna-13B  & 224&129M/1.2M & -- & 49.5 & 63.1 & 50.7 & 78.9 & 1212.8 & -- & 25.6 \\ 
Shikra~\citep{chen2023shikra} & Vicuna-13B  & 224&600K/5.5M & 77.4 & -- & -- & -- & -- & -- & 58.8 & -- \\ 
Otter~\citep{Li2023OtterAM} & LLaMA-7B & 224& -&-&-&-&-&-& 1292.3 & 48.3 & 24.6\\
mPLUG-Owl~\citep{ye2023mplug} & LLaMA-7B  & 224&2.1M/102K & -&-&-&-&-&967.3&49.4&-\\
IDEFICS-9B~\citep{idefics} & LLaMA-7B & 224&353M/1M & 50.9 & 38.4 & -- & 25.9 & -- & -- & 48.2 & -- \\ 
IDEFICS-80B~\citep{idefics} & LLaMA-65B &224&353M/1M & 60.0 & 45.2 & -- & 30.9 & -- & -- & 54.5 & -- \\
Qwen-VL~\citep{bai2023qwen} & Qwen-7B  &448&1.4B/50M & \textbf{78.8} & 59.3 & 67.1 & \textbf{63.8} & -- & -- & 38.2 & -- \\ 
Qwen-VL-Chat~\citep{bai2023qwen}  & Qwen-7B & 448&1.4B/50M & 78.2 & 57.5 & 68.2 & 61.5 & -- & 1487.5 & 60.6 & -- \\ 
LLaVA-1.5~\citep{liu2023improvedllava} & Vicuna-7B & 336&558K/665K&78.5&62.0&66.8&58.2&\textbf{85.9}&\textbf{1510.7}&64.3&30.5\\
\midrule
LLaVA-Phi~\citep{Zhu2024LLaVAPhiEM} & Phi2-2.7B &336&558K/665K& 71.4 & - &\textbf{68.4} & 48.6 & 85.0 & 1335.1 & {59.8} & 28.9\\
MobileVLM-3B~\citep{Chu2023MobileVLMA} & LLaMA-2.7B& 336 &558K/665K& - & {59.0} & 61.2 &47.5 & 84.9 & 1288.9 & 59.6 & -\\
VL-Mamba~\citep{Qiao2024VLMambaES}& Mamba{-2.8B}& 224 &558K/665K&76.6 & 56.2 & 65.4 & {48.9} &{84.4} &{1369.6}& 57.0& {32.6}\\
VisualRWKV-Base~\citep{Peng2024EagleAF}& RWKV5-1.6B& 336 &558K/665K& 51.1 & 48.1 & 41.9 & 35.2 &73.1 & - & - & - \\
\midrule
\textbf{VisualRWKV} & RWKV6-1.6B&336&558K/665K & 69.4 & 55.2 & 59.1 & 43.6 & 83.2&1204.9&55.8&53.2\\
\textbf{VisualRWKV} & RWKV6-3B &336&558K/665K& 71.5 & 59.6 & 65.3& 48.7 &83.1&1369.2&59.5&56.3\\
\textbf{VisualRWKV} & RWKV6-7B &336&558K/665K& 75.8 & \textbf{64.3} & 68.2 & 51.0 &84.7&1387.8&\textbf{65.8}&\textbf{63.7}\\
\bottomrule
\end{tabular}
}
\caption{\textbf{Comparison with SoTA methods on 8 benchmarks.} 
Due to space constraints, benchmark names are abbreviated. VQA~\citep{goyal2017vqav2}; GQA~\citep{hudson2019gqa}; SQA: ScienceQA-IMG~\citep{lu2022learn}; TQA: TextVQA~\citep{singh2019textvqa}; POPE~\citep{li2023pope}; MME~\citep{fu2023mme}; MMB: MMBench~\citep{Liu2023MMBenchIY}; MMB-cn: MMBench-CN~\citep{Liu2023MMBenchIY}.
PT and IT denote the quantity of samples involved in the pre-training and instruction-tuning phases. "Res." stands for "Resolution.
}
\label{tab:results}
\end{table*}

\subsubsection{Gain Analysis on Different Benchmarks}
VisualRWKV excels in academic benchmarks like VQA, GQA, and SQA, where both the questions and answers are short texts. 
The model faces no fundamental obstacles in handling such tasks, leading to significant performance improvements. 
As a result, VisualRWKV achieves results that are comparable to, and even surpass, the Transformer-based LLaVA-1.5 on these benchmarks.

Although VisualRWKV shows notable improvements on the TextVQA (TQA) benchmark, it still lags behind LLaVA-1.5 in this task (51.0 vs. 58.2). 
TextVQA requires recalling information from images, which is similar to the Multi-Query Associative Recall (MQAR) task~\cite{arora2023zoology}, which is often a limitation for RNN-like architectures.
However, our latest work, VisualRWKV-HD/UHD~\cite{Li2024VisualRWKVHDAU}, has shown that higher resolution and better-quality image features can significantly alleviate these limitations.

\subsection{Ablation Study}
\subsubsection{Ablation on Data-dependent Recurrence}
To verify the effectiveness of data-dependent recurrence described in Section \ref{sec:data-dep}, we conducted a rigorous ablation study, ensuring that the model size, training data, environment, and all hyperparameters were strictly consistent.
As depicted in Table \ref{tab:scaling_ablation}, the outcomes demonstrate significant enhancements in the data-dependent VisualRWKV across the four monitored benchmarks, affirming that data-dependence is essential for the success of linear RNN-type models in the VLM domain.

\subsubsection{Ablation on Prompting Method}

As shown in Table \ref{tab:prompt_res}, among the three prompting approaches, the sandwich prompt outperforms the others, followed by the image-first prompt, with the image-last prompt being the least effective. The effectiveness of the sandwich prompt is attributed to its ability to allow the model to review the instructions before engaging with the image, enabling a more targeted extraction of information and enhancing the conditional aspects of image information retrieval.

However, simply placing the instructions before the image is insufficient. The image-last prompt performs poorly because linear RNN models tend to forget the instructions after processing the image, making it necessary to repeat the instructions for better results. Additionally, our research shows that the sandwich prompt can effectively mitigate information loss even with a reduced number of image tokens, maintaining robust performance. Further experimental results and analyses can be found in Appendix \ref{sec:app_prompt}.

\begin{table}[h]
\centering
\resizebox{\columnwidth}{!}{%
\begin{tabular}{@{}lcccccc@{}}
\toprule
\textbf{Method} & \textbf{Size} & \textbf{Prompt} & \textbf{VQA} & \textbf{SQA} & \textbf{TQA} & \textbf{GQA} \\ \midrule
\textbf{VisualRWKV-Base} & 7B & First & 67.93 & \textbf{65.59} & 47.13 & 48.52 \\
\textbf{VisualRWKV-Base} & 7B & Last & 63.07 & 57.66 & 48.52 & 44.19 \\
\textbf{VisualRWKV-Base} & 7B & Sandwich & \textbf{69.71} & 65.20 & \textbf{50.25} & \textbf{50.50} \\ \bottomrule
\end{tabular}%
}
\caption{
Results for three prompting method.
}
\label{tab:prompt_res}
\end{table}

\subsubsection{Ablation on Scanning Method}

We compared three image scanning mechanisms: Uni-directional scanning (UniDir), Bi-directional scanning (BiDir), and Multi-directional scanning (MultiDir).
As shown in Table \ref{tab:scan_res}, UniDir performs the worst because it is inherently unsuitable for 2D visual information.
BiDir and MultiDir show comparable outcomes across various benchmark assessments, but BiDir outperforms in the majority, highlighting its strength in handling 2D visual information for multimodal learning tasks.

The image scanning techniques are applied during both training and inference, and it is essential to maintain train-test consistency. We have made simple attempts to rearrange the order of layers with different directions, but the performance was not robust. Specific layers have already been specialized to process image information from particular directions.

\begin{table}[h]
\centering
\resizebox{\columnwidth}{!}{%
\begin{tabular}{@{}lccllll@{}}
\toprule
\textbf{Method} & \textbf{Size} & \textbf{Scanning} & \multicolumn{1}{c}{\textbf{VQA}} & \multicolumn{1}{c}{\textbf{SQA}} & \multicolumn{1}{c}{\textbf{TQA}} & \multicolumn{1}{c}{\textbf{GQA}} \\ \midrule
\textbf{VisualRWKV-Base} & 1.6B & UniDir & 51.03 & 41.94 & 35.19 & 48.09 \\
\textbf{VisualRWKV-Base} & 1.6B & BiDir & 65.62 & \textbf{47.30} & \textbf{37.13} & 48.60 \\
\textbf{VisualRWKV-Base} & 1.6B & MultiDir & \textbf{66.04} & 44.03 & 35.84 & \textbf{49.95} \\ \midrule
\textbf{VisualRWKV} & 1.6B & BiDir & \textbf{69.26} & \textbf{57.61} & \textbf{43.17} & \textbf{54.85} \\
\textbf{VisualRWKV} & 1.6B & MultiDir & 69.20 & 57.31 & 42.97 & 54.63 \\ \bottomrule
\end{tabular}%
}
\caption{
Results for three scanning methods.
}
\label{tab:scan_res}
\end{table}

\subsubsection{Ablation on Learning Rate}
As shown in Table \ref{tab:scaling_ablation}, correct learning rate is crucial for the performance of benchmarks.  
Table \ref{tab:learning_rate} shows a comparison of our model with different learning rate.
From the Table, it can be observed that a higher initial learning rate has a significant impact on the model's performance.
Our hypothesis is that
the substantial divergence in tasks from the textual to the visual domain necessitates a higher learning rate to facilitate the model's adaptation.

It has been observed that there is a substantial discrepancy between the optimal learning rates of VisualRWKV and LLaVA\citep{liu2023improvedllava}, with the optimal initial learning rate for LLaVA-1.5-7B being $2e^{-5}$ and for VisualRWKV-7B being $4e^{-5}$. This caused considerable difficulties in our work at the beginning and also confirmed the significant divergence between the RWKV architecture and the Transformer architecture.

\subsection{Efficiency Analysis}
As shown in Figure \ref{fig:teaser}, we compared the inference speed and GPU memory consumption directly with LLaVA-1.5 of the same parameter size.
VisualRWKV has a constant single token inference speed, while the inference speed of a single token in LLaVA-1.5 slows down as more tokens are generated.
On the other hand, VisualRWKV has a constant GPU memory consumption, while the memory consumption of LLaVA-1.5 increases linearly. 
In practice, compared to LLaVA-1.5, VisualRWKV has a speed advantage of 3.98 times and can save 54\% of the GPU memory when reaching an inference length of 24576 tokens.
Since VisualRWKV retains a fixed state size throughout inference, GPU memory usage remains nearly constant, which is illustrated as a straight line in Figure~\ref{fig:teaser}(c).

\subsection{Text-only Capability}
According to \citet{lin2023vila}, LLMs face the issue of degraded text capabilities after visual instruction tuning.
As shown in Table \ref{tab:text_capability}, no degradation of text abilities was observed in VisualRWKV.
Conversely, enhancements in performance were noted across various text-only English datasets, which we credit to the integration of a large set of English samples in our fine-tuning dataset.
Furthermore, it was observed that VisualRWKV did not face text ability degradation across multiple languages, as shown in Table \ref{tab:text_capability}. The capabilities were fundamentally aligned with those of the text-only RWKV. 
This may be due to the incorporation of the multilingual ShareGPT4. More details about text-only capability can be found in Appendix \ref{sec:app-text-only}.

\begin{table}[h]
\centering
\resizebox{\columnwidth}{!}{%
\begin{tabular}{@{}lcccc@{}}
\toprule
\textbf{Method} & \textbf{Size} & \textbf{LAMBADA} & \textbf{English} & \textbf{MultiLang} \\ 
  & & ppl &avg\% & avg\% \\ \midrule
\textbf{RWKV} & 1.6B & 4.63 & 59.82 & 59.97 \\
\textbf{VisualRWKV} & 1.6B & 4.15 & 61.01 & 59.83 \\ \bottomrule
\end{tabular}%
}
\caption{
Results for text-only capability: The English score is the average of 10 English benchmarks, while the Multilingual score is the average of 4 Multilingual benchmarks.
}
\label{tab:text_capability}
\end{table}

Besides the results previously stated, we also compared the outcomes of single-stage and two-stage training approaches; conducted ablation studies on the method of cross-entropy loss reduction; assessed the influence of Weight Decay on the model; and explored a basic hybrid model known as VisualRWKV Hybrid. Due to space limitations, we have included these contents in the Appendix.

\section{Conclusions}
In this paper, we introduce for the first time VisualRWKV, which explores the construction of a visual language model using the linear RNN model RWKV.
VisualRWKV incorporates three innovative designs: data-dependent recurrence to enhance the model's information extraction capabilities, sandwich prompt for better conditioning, and bidirectional scanning for more effective extraction of 2D visual information.
We conducted extensive experiments on eight multimodal benchmarks and achieved comparable performance with some of the most advanced VLMs; we also carried out ablation studies to evaluate the effectiveness of data-dependent recurrence, prompting methods, and various scanning mechanisms.
The results validate the effectiveness of our proposed model and demonstrate the potential of applying RNNs to VLMs.

\section*{Limitations}
Despite the encouraging results achieved by VisualRWKV, several limitations must be acknowledged.
Firstly, due to the lack of data following such instructions and the limited context length, VisualRWKV is currently unable to process multiple images.
Secondly, although VisualRWKV shows good performance on academic datasets, its ability to handle certain tasks, such as TextVQA, may be constrained by the limitations in the recall ability of efficient language models \citep{arora2023zoology}. These constraints can potentially be mitigated by further architectural improvements.
Lastly, to maintain consistency with LLaVA-1.5, this study did not investigate the effects of the choice of vision encoder or the quality of training data on VisualRWKV. In the future, we aim to explore more advanced visual encoders and utilize higher-quality training data to further enhance its performance.

\paragraph{Risks} Although VisualRWKV significantly reduces the occurrence of hallucinations, it can still generate hallucinations and occasionally disseminate misinformation. Therefore, its application in critical fields, such as the medical industry, should be approached with great caution.

\section*{Acknowledgments}
Thanks to Peng Bo, the author of RWKV, for participating in the discussion and providing valuable suggestions for modifications. 

\bibliography{custom}

\newpage
\appendix
\onecolumn

\textbf{Supplementary Material for VisualRWKV: Exploring Recurrent Neural Networks for Visual Language Models}

\section{Data and Hyperparameters}
\label{sec:app_data_Hyperparameters}
\paragraph{Training Data}
The data used in this study is strictly aligned with LLaVA-1.5.
The training of VisualRWKV is composed of two phases: (1) Feature Alignment Phase: Utilizing our 558K subset from the LAION-CC-SBU dataset, we link a pretrained, frozen vision encoder to a frozen Large Language Model (LLM);
(2) Visual Instruction Tuning Phase: We employ 150K of GPT-generated multimodal instruction-following datasets, supplemented by approximately 515K Visual Question Answering (VQA) datasets from academically oriented tasks \citep{okvqa,singh2019textvqa,hudson2019gqa,goyal2017vqav2}, to instruct the model in adhering to multimodal directives.
For more details, one can refer to the paper on LLaVA-1.5 \citep{liu2023improvedllava}.
All the data used in this paper are consistent with their intended use.
We carefully identified and handled all personally identifiable information and offensive content. We started with automated screening to flag sensitive data, followed by manual review for precision. Anonymization methods like data masking and pseudonymization were applied to protect sensitive information. Strict data protection protocols were followed throughout.

\paragraph{Evaluation Benchmarks}
Additional details on Benchmarks are provided here.
The VQA-v2 reports its metrics based on the test-dev split.
Similarly, GQA's metrics are on the test-dev split.
The metrics for TextVQA are reported on the validation set.
ScienceQA's metrics are based on the test set.
POPE's metrics are also reported on the test set.
The MMBench metrics are reported on the development set.
MME has a unique test-set, thus there is no ambiguity.

\paragraph{Data Language}
Firstly, the training data includes academic Visual Question Answering (VQA) datasets and ShareGPT data.
The primary language of the VQA academic datasets is English, while the ShareGPT data is multilingual, encompassing mainstream languages, but derived from contributions by users worldwide, it is not feasible to count the total number of languages.
Among the evaluation benchmarks, MMBench-cn is the only Chinese dataset; the rest are English datasets.
Concurrently, we evaluated the model's text-only capabilities in multiple languages, with the specific languages detailed in Appendix \ref{sec:app-text-only}.

\paragraph{Hyperparameters}
The hyperparameters here were used for the training of a range of VisualRWKV models, from 1.6B to 7B parameters, as illustrated in Table \ref{tab:results}. We show the training hyperparameters for both first-stage vision-language alignment pretraining and the second-stage visual instruction tuning in Table~\ref{tab:hyperparameter}.

\begin{table}[h!]
\centering
\resizebox{\textwidth}{!}{
\begin{tabular}{l|c ccccc}
\toprule
Hyperparameter & 1.6B-Pretrain& 1.6B-Finetune & 3B-Pretrain& 3B-Finetune& 7B-Pretrain&7B-Finetune\\
\midrule
batch size & 256 & 128  & 256 & 128  & 256 &128  \\
lr init& 1e-3 & 6e-5& 1e-3 
& 5e-5& 1e-3 
&4e-5\\
 lr end& 1e-5 & 1.5e-5& 1e-5 & 1.25e-5& 1e-5 &1e-5\\
lr schedule & cosine decay&cosine decay&cosine decay&cosine decay&cosine decay&cosine decay\\
lr warmup ratio & 0&0&0&0&0&0\\
weight decay & 0&0&0&0&0&0\\
epoch & 1& 2&  1& 2& 1&2\\
optimizer & AdamW&AdamW&AdamW&AdamW&AdamW&AdamW\\
 DeepSpeed stage & 1& 1 & 1& 1& 1&2\\
\bottomrule
\end{tabular}
}
\caption{
\textbf{Hyperparameters} of \newshortname{}.
}
\label{tab:hyperparameter}
\end{table}

\paragraph{Licenses}
VisualRWKV is licensed under the Apache-2.0 license. The RWKV language model is also under the Apache-2.0 license. The LLaVA model is licensed under the Apache-2.0 license. The VQA-v2 dataset is licensed under the Commons Attribution 4.0 International License. MMBench is licensed under the Apache-2.0 license. TextVQA data is available under the CC BY 4.0 license. ScienceQA is licensed under the MIT License, and POPE is also under the MIT license.

\section{Model and Computation}

\paragraph{LLM Model}
The LLM foundation model is primarily based on two families: the RWKV-5 series\footnote{\url{https://huggingface.co/BlinkDL/rwkv-5-world}} and the RWKV-6 series\footnote{\url{https://huggingface.co/BlinkDL/rwkv-6-world}}.
Both the RWKV-5 and RWKV-6 series consist of models with 1.6 billion, 3 billion, and 7 billion parameters respectively.
In this research, the RWKV-5 series is mainly applied in the VisualRWKV-Base, and the RWKV-6 series acts as the LLM backbone for VisualRWKV.

\paragraph{Model Size}
The vision encoders utilized in this paper are based on the CLIP-L model, which features 0.3 billion parameters. In contrast, the RWKV models vary in size: the RWKV 7B has 7.6 billion parameters, the RWKV 1.6B has 1.6 billion parameters, and the RWKV 3B has 3.1 billion parameters. Consequently, the VisualRWKV variants have different total parameter counts: the VisualRWKV 1.6B encompasses 1.9 billion parameters, the VisualRWKV 3B includes 3.4 billion parameters, and the VisualRWKV 7B comprises 7.9 billion parameters.

\paragraph{Computing Infrastructure}
A range of computational resources were employed in the study. The standard training and benchmark evaluation were conducted using 8 NVIDIA A100-80GB GPUs. The VisualRWKV 7B model is trained with 6 A100 GPUs due to insufficient memory capacity with 8 GPUs.
For the efficiency analysis, a GPU with L20-48GB of memory was employed.

\paragraph{Computing Budget}
Training an epoch of VisualRWKV 1.6B with 8 A100 GPUs takes 6.7 hours, equivalent to 53.6 GPU hours;
Training an epoch of VisualRWKV 3B with 8 A100 GPUs takes 11.3 hours, equivalent to 90.4 GPU hours;
Training an epoch of VisualRWKV 7B with 6 A100 GPUs takes 26.5 hours, equivalent to 159 GPU hours.

\paragraph{Packages Version}
The main experimental environment for this study is the NVIDIA PyTorch NGC Container (23.07-py3) with lightning1.9.5 and deepspeed0.12.6. For updates, please refer to our codebase (currently anonymized, will be released later).

\section{Single-Stage Training vs. Two-Stage Training}

The research conducted by ~\citet{Karamcheti2024PrismaticVI}, suggests that including a distinct projector pretraining phase may not be essential. Their study indicates that a single-stage training process can lead to improved performance outcomes. Omission of the pretraining phase results in a significant cost reduction of about 20 to 25 percent and avoids the need for stage-specific data collection. 

\begin{figure}[h!]
    \centering
    \includegraphics[width=\linewidth]{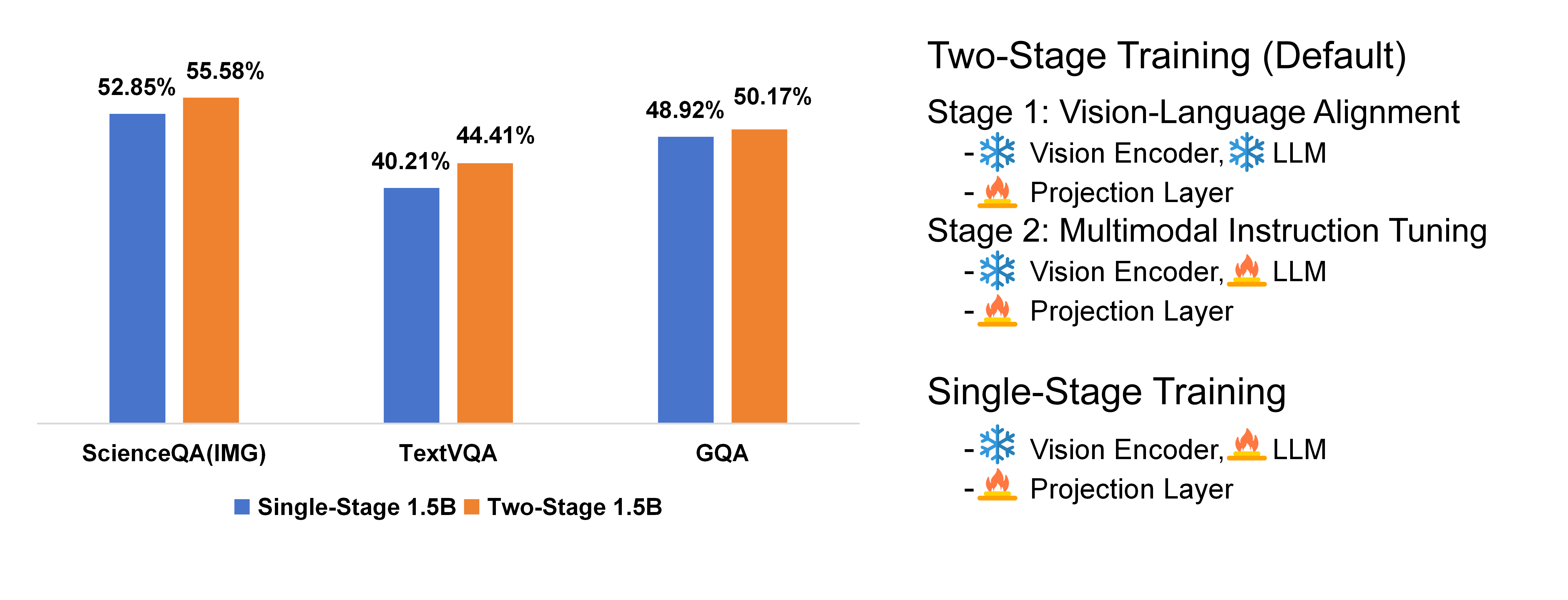}
    \caption{\textbf{Single-Stage Training vs. Two-Stage Training.} We conducted a comparative analysis between the two-stage training and single-stage training, with the latter omitting the vision-language alignment phase. Our findings reveal that the single-stage training yields inferior performance outcomes. This suggests that the vision-language alignment, integral to the two-stage training, significantly contributes to enhanced performance.}
    \label{fig:single-stage}
\end{figure}

To validate these insights, we conducted a series of experiments using the VisualRWKV framework. The results, as illustrated in Figure \ref{fig:single-stage}, show that the two-stage training outperforms single-stage training, indicating that the two-stage approach is still very necessary. The different results associated with single-stage training could be due to the diverse training setups used by various researchers. Given these results, we have made a strategic decision to \textbf{adopt a two-stage training protocol for all subsequent experiments in this paper}.

\section{Influence of Cross-Entropy Loss Reduction}

In the experiment, we found that using zero1 for training with a batch size of 1 and gradient accumulation of 16; and using zero2 for training with a batch size of 1 and gradient accumulation of 1; These two settings are not equivalent, with different losses, leading to significantly disparate outcomes for the final model. Therefore, we conducted an in-depth analysis and study. 

For illustrative purposes, consider a simple thought experiment with four samples: the first sample consists of 100 tokens, the second of 200 tokens, the third of 300 tokens, and the fourth of 400 tokens. Consequently, the total length sums up to 1000 tokens. When these samples are batched together(batch size of 4 and gradient accumulation of 1), each token is normalized by a factor of 1000. We refer this process as batch-level reduction. Please note that the batch-level reduction is highly dependent on the batch size. As the batch size varies, the total batch length by which each token's loss is divided can differ significantly.

An alternative approach, termed sample-level reduction, normalizes each sample by its length. This sample-level reduction is independent of the batch size and introduces a different loss re-weighting compared to batch-level reduction. Continuing our thought experiment, we apply sample-level reduction with a batch size of 1 and gradient accumulation of 4. The first sample undergoes a sequential division by 100 (its length) and then by 4, culminating in an effective division by 400. The second sample is adjusted by a factor of 800, the third by 1200, and the fourth by 1600. This scaling mechanism inherently leads to a larger loss for shorter texts and a smaller loss for longer texts compared to batch-level reduction.

Our findings underscore the importance of accurate reduction and loss re-weighting for the performance of certain downstream tasks. Table \ref{tab:loss_reduction} presents a comparative analysis between our model's performance under batch-level and sample-level reduction. Notably, we have found that using sample-level reduction yields better results on 5 benchmarks. In contrast, batch-level reduction performs better on 2 benchmarks. Among them, sample-level reduction significantly outperforms on the ScienceQA benchmark. On the MME benchmark, batch-level reduction takes the lead. After an in-depth investigation, we discovered that the score in the Celebrity domain within MME has significantly improved, while other domains show varying degrees of success.

\begin{table}[h]
\centering
\resizebox{0.8\textwidth}{!}{%
\begin{tabular}{@{}lccccccc@{}}
\toprule
\textbf{Reduction} & \textbf{VQAv2} & \textbf{ScienceQA} & \textbf{TextVQA} & \textbf{GQA} & \textbf{VizWiz} & \textbf{MME} & \textbf{POPE} \\ \midrule
\textbf{Sample-Level} & \textbf{67.54} & \textbf{56.62\%} & \textbf{42.18\%} & \textbf{52.82\%} & 26.03 & 1111.66 & \textbf{0.82} \\
\textbf{Batch-Level} & 66.85 & 47.94\% & 41.79\% & 52.56\% & \textbf{27.02} & \textbf{1173.42} & 0.79 \\ \bottomrule
\end{tabular}%
}
\caption{Study comparing batch-level reduction and sample-level reduction across 7 Visual Language benchmarks. Loss reduction method is crucial for performance. Model used here is VisualRWKV 1.6B.}
\label{tab:loss_reduction}
\end{table}

Furthermore, we conducted a comparison of the textual abilities resulting from sample-level and batch-level reduction, as shown in Table \ref{tab:reduction_text_ability}. It was observed that sample-level training exhibited superior English capabilities, whereas the batch-level training demonstrated enhanced multilingual abilities.
This is due to the higher loss weight assigned to the multilingual long texts of ShareGPT4 data in the batch-level training.

In general, we consider sample-level reduction to be the better approach.
On one hand, the performance is better, whether in visual-linguistic abilities or pure textual capabilities.
On the other hand, sample-level reduction is invariant to batch size. When the sample-level reduction-based training protocol is migrated across various GPUs, it does not suffer from inconsistencies due to batch size variations, which could otherwise lead to divergent outcomes.

\begin{table}[h]
\centering
\resizebox{0.65\textwidth}{!}{%
\begin{tabular}{@{}lccc@{}}
\toprule
\textbf{Reduction} & \textbf{LAMBADA(ppl)} & \textbf{English(avg\%)} & \textbf{MultiLang(avg\%)} \\ \midrule
\textbf{Batch-Level} & 4.499 & 59.89 & \textbf{59.97} \\
\textbf{Sample-Level} & \textbf{4.145} & \textbf{61.01} & 59.84 \\ \bottomrule
\end{tabular}
}
\caption{Study comparing batch-level reduction and sample-level reduction across language benchmarks. Model used here is VisualRWKV 1.6B.}
\label{tab:reduction_text_ability}
\end{table}

\section{Further details on the Prompting Method}
\label{sec:app_prompt}

In this section, we will further discuss three types of prompt methods.
As shown in Table \ref{tab:app-prompt}, we found that as the number of image tokens decreases, the effectiveness of the image first prompt and sandwich prompt also monotonically decreases, which is intuitively expected as fewer image tokens contain less pictorial information.
Nonetheless, the image last prompt does not exhibit a strictly decreasing trend; it initially increases and subsequently decreases, achieving optimal performance at the point of 145 image tokens.
The effect is especially evident in scenarios of train-test mismatch.
We term this the information barrier formed by image tokens, which hinders the model's information transfer.

An additional observation indicates that the sandwich prompt is capable of mitigating information loss, sustaining good performance even with a limited number of image tokens.
In contrast, the other two types of prompt methods fail to achieve this.

\begin{table}[h!]
\centering
\resizebox{0.8\textwidth}{!}{%
\begin{tabular}{@{}lccclll@{}}
\toprule
\textbf{Method} & \textbf{Size} & \textbf{Prompt} & \multicolumn{1}{l}{\textbf{Image Tokens}} & \textbf{ScienceQA} & \multicolumn{1}{c}{\textbf{TextVQA}} & \multicolumn{1}{c}{\textbf{GQA}} \\ \midrule
\multirow{8}{*}{VisualRWKV-Base} & \multirow{8}{*}{7B} & \multirow{8}{*}{First} & 577 & 65.59\% & 47.13\% & 48.52\% \\
 &  &  & 145 & 64.14\% & 42.91\% & 45.99\% \\
 &  &  & 65 & 64.01\% & 40.67\% & 44.08\% \\
 &  &  & 37 & 62.87\% & 39.90\% & 43.44\% \\
 &  &  & 17 & 61.23\% & 39.96\% & 43.31\% \\
 &  &  & 10 & 60.29\% & 39.65\% & 43.23\% \\
 &  &  & 5 & 59.35\% & 39.80\% & 43.16\% \\
 &  &  & 1 & 57.11\% & 39.34\% & 43.53\% \\ \midrule
\multirow{8}{*}{VisualRWKV-Base} & \multirow{8}{*}{7B} & \multirow{8}{*}{Last} & 577 & 57.66\% & 48.52\% & 44.19\% \\
 &  &  & 145 & 58.75\% & 45.29\% & 42.93\% \\
 &  &  & 65 & 56.07\% & 43.89\% & 42.38\% \\
 &  &  & 37 & 53.35\% & 43.03\% & 42.07\% \\
 &  &  & 17 & 50.37\% & 42.50\% & 42.03\% \\
 &  &  & 10 & 50.72\% & 42.18\% & 42.10\% \\
 &  &  & 5 & 49.23\% & 41.20\% & 41.80\% \\
 &  &  & 1 & 50.67\% & 41.19\% & 41.93\% \\ \midrule
\multirow{8}{*}{VisualRWKV-Base} & \multirow{8}{*}{7B} & \multirow{8}{*}{Sandwich} & 577 & 65.20\% & 50.25\% & 50.50\% \\
 &  &  & 145 & 64.90\% & 46.38\% & 47.47\% \\
 &  &  & 65 & 64.40\% & 44.58\% & 45.09\% \\
 &  &  & 37 & 64.11\% & 44.01\% & 44.78\% \\
 &  &  & 17 & 63.86\% & 43.61\% & 44.57\% \\
 &  &  & 10 & 63.26\% & 43.27\% & 44.37\% \\
 &  &  & 5 & 62.87\% & 43.03\% & 44.08\% \\
 &  &  & 1 & 60.34\% & 41.72\% & 36.09\% \\ \bottomrule
\end{tabular}%
}
\caption{Full Results for three prompting method.}
\label{tab:app-prompt}
\end{table}

\section{Study on Learning Rate}

In this section, We will explore the effect of learning rates on VisualRWKV. 
Setting different initial learning rates and using a cosine learning rate scheduler, the performance of the model on multiple benchmarks is shown in the Table \ref{tab:learning_rate}.

\begin{table}[h]
\centering
\resizebox{0.9\textwidth}{!}{%
\begin{tabular}{@{}lllcccccc@{}}
\toprule
\textbf{Method} &
  \multicolumn{1}{c}{\textbf{Size}} &
  \multicolumn{1}{c}{\textbf{Learning Rate}} &
  \multicolumn{1}{c}{\textbf{VQAv2}} &
  \multicolumn{1}{c}{\textbf{ScienceQA}} &
  \multicolumn{1}{c}{\textbf{TextVQA}} &
  \multicolumn{1}{c}{\textbf{GQA}} &
  \multicolumn{1}{c}{\textbf{MME}} \\ \midrule
\textbf{VisualRWKV} & 1.6B & 2e-5 to 2e-5   & 66.85 & 57.51 & 41.85 & 52.07 & 1080.77 \\
\textbf{VisualRWKV} & 1.6B & 3e-5 to 1e-5   & 67.25 & 53.40 & 41.84 & 52.49  & 1115.70 \\
\textbf{VisualRWKV} & 1.6B & 3e-5 to 1.5e-5 & 67.54 & 56.62 & 42.18 & 52.82 & 1111.66 \\
\textbf{VisualRWKV} & 1.6B & 4e-5 to 1.5e-5 & 68.51 & 55.68 & 43.73 & 54.31 & 1151.20 \\ 
\textbf{VisualRWKV} & 1.6B & 5e-5 to 1.5e-5 & 69.26 & 57.61 & 43.17 & 54.85 & 1208.96 \\ 
\textbf{VisualRWKV} & 1.6B & \textbf{6e-5 to 1.5e-5} & 69.42 & 59.05 & 43.57 & 55.23 & 1204.90 \\ 
\textbf{VisualRWKV} & 1.6B & 1e-4 to 1.5e-5 & 70.02 & 55.58 & 42.24 & 55.72 & 1212.52 \\ 
\textbf{VisualRWKV} & 1.6B & 1.5e-4 to 1.5e-5 & 68.89 & 55.63 & 41.90 & 54.09 & 1249.51 \\ 
\midrule
\textbf{VisualRWKV} & 3B & 4e-5 to 1e-5 & 68.65 & 65.99 & 48.46 & 54.40 & 1323.18 \\ 
\textbf{VisualRWKV} & 3B & \textbf{5e-5 to 1.25e-5} & 71.52 & 65.34 & 48.68 & 59.56 & 1369.19 \\ 
\midrule
\textbf{VisualRWKV} & 7B & 2e-5 to 2e-5 & 68.31 & 68.91 & 50.09 & 52.80 & 1340.44 \\ 
\textbf{VisualRWKV} & 7B & \textbf{4e-5 to 1e-5} & 75.82 & 68.22 & 51.01 & 64.27 & 1387.75 \\ 
\bottomrule
\end{tabular}%
}
\caption{Impact of Learning Rate on the Performance of the VisualRWKV on 5 benchmarks.}
\label{tab:learning_rate}
\end{table}

\section{Improvement on Text-only Capability}
\label{sec:app-text-only}
In this section, you can find full results on text-only capability, as shown in the Table \ref{tab:english_performance} and Table \ref{tab:multilang_performance}.

\begin{table}[h]
\resizebox{\textwidth}{!}{%
\begin{tabular}{@{}lccccccccccccc@{}}
\toprule
\textbf{Method} & \textbf{Size} &\textbf{LBD} & \textbf{Eng} & \textbf{LAM} & \textbf{PIQA} & \textbf{SC16} & \textbf{HSW} & \textbf{WG} & \textbf{ARC-C} & \textbf{ARC-E} & \textbf{HQA} & \textbf{OBQA} & \textbf{SCIQ} \\
& & ppl & avg\% & acc & acc & acc & acc-n & acc & acc-n & acc & acc-n & acc-n & acc \\  \midrule
\textbf{RWKV} & 1.6B & 4.63 & 59.82 & 67.39 & 74.37 & 74.50 & 61.06 & 60.93 & 33.70 & 64.18 & 35.22 & 37.4 & 89.40 \\
\textbf{VisualRWKV} &1.6B & 4.15 & \textbf{61.01} & 67.64 & 73.44 & 75.09 & 61.50 & 61.95 & 38.31 & 67.88 & 36.46 & 38.0 & 89.80 \\ \bottomrule

\end{tabular}%
}
\caption{The table lists the English performance metrics for various benchmarks: LBD (LAMBADA), PIQA, SC16 (StoryCloze16), HSW (Hellaswag), WG (WinoGrande), ARC-C (arc\_challenge), ARC-E (arc\_easy), HQA (headQA\_en), OBQA (openbookQA), SCIQ. Metric units are ppl (perplexcity), acc (accuracy) and acc-n (normalized accuracy).}
\label{tab:english_performance}
\end{table}

For multilingual evaluations, we assess LAMBADA in English, French, German, Italian, and Spanish. We evaluate StoryCloze as per \citep{Lin2021FewshotLW} in Arabic, English, Spanish, Basque, Hindi, Indonesian, Burmese, Russian, Swahili, Telugu, and Chinese. COPA is evaluated in Estonian, Haitian Creole, Indonesian, Italian, Cusco-Collao Quechua, Kiswahili, Tamil, Thai, Turkish, Vietnamese, and Chinese, following \citep{Ponti2020XCOPAAM}. We also evaluate multilingual WinoGrande in English, French, Japanese, Portuguese, Russian, and Chinese, as demonstrated in \citep{tikhonov2021heads, muennighoff2022crosslingual}.

\begin{table}[h]
\centering
\resizebox{0.6\textwidth}{!}{%
\begin{tabular}{@{}lcccccc@{}}
\toprule
\textbf{Method} & \textbf{Size} & \textbf{MultiLang} & \textbf{xLBD} & \textbf{xSC} & \textbf{xWG} & \textbf{xCOPA} \\
  & & avg\% & acc & acc & acc & acc \\ \midrule
\textbf{RWKV} & 1.6B & 59.97 & 47.17 & 58.24 & 76.46 & 58.03 \\
\textbf{VisualRWKV} & 1.6B & 59.83 & 46.73 & 58.90 & 75.07 & 58.65 \\ \bottomrule
\end{tabular}
}
\caption{The table lists the Multi-Language performance metrics for various benchmarks: xLBD (Multilingual LAMBADA), xSC (Multilingual StoryCloze), xWG (Multilingual WinoGrande), xCOPA (Multilingual COPA).}
\label{tab:multilang_performance}
\end{table}

\section{Study on Weight Decay}

\begin{table}[h]
\centering
\resizebox{0.9\textwidth}{!}{%
\begin{tabular}{@{}lcccccccc@{}}
\toprule
\textbf{Model} & \textbf{Weight Decay} & \textbf{Learning Rate} & \textbf{VQA} & \textbf{SQA} & \textbf{TQA} & \textbf{GQA} & \textbf{VizWiz} & \textbf{MME} \\ \midrule
\textbf{VisualRWKV 1.6B} & 0 & 6e-5 to 1.5e-5 & \textbf{69.42} & 59.05 & \textbf{43.57} & \textbf{55.23} & \textbf{29.84} & \textbf{1204.90} \\
\textbf{VisualRWKV 1.6B} & 0.1 & 6e-5 to 1.5e-5 & 68.48 & 58.85\% & 41.58 & 54.34 & 28.05 & 1173.03 \\
\textbf{VisualRWKV 1.6B} & 0.01 & 6e-5 to 1.5e-5 & 68.53 & \textbf{59.40\%} & 42.24 & 54.24 & 27.86 & 1154.52 \\ \bottomrule
\end{tabular}%
}
\caption{Impact of Weight Decay on the Performance of the VisualRWKV on 6 benchmarks.}
\label{tab:weight_decay}
\end{table}

Having established the best learning rate, we conducted additional investigations into weight decay.
Weight decay was imposed solely on the model's linear layers.
The Table \ref{tab:weight_decay} illustrates that, currently, the best outcomes are achieved without weight decay.
The role of weight decay is complex and may require further exploration in the future.

\section{VisualRWKV Hybrid}
\begin{figure*}[hbt]
    \centering
    \includegraphics[width=0.68\linewidth]{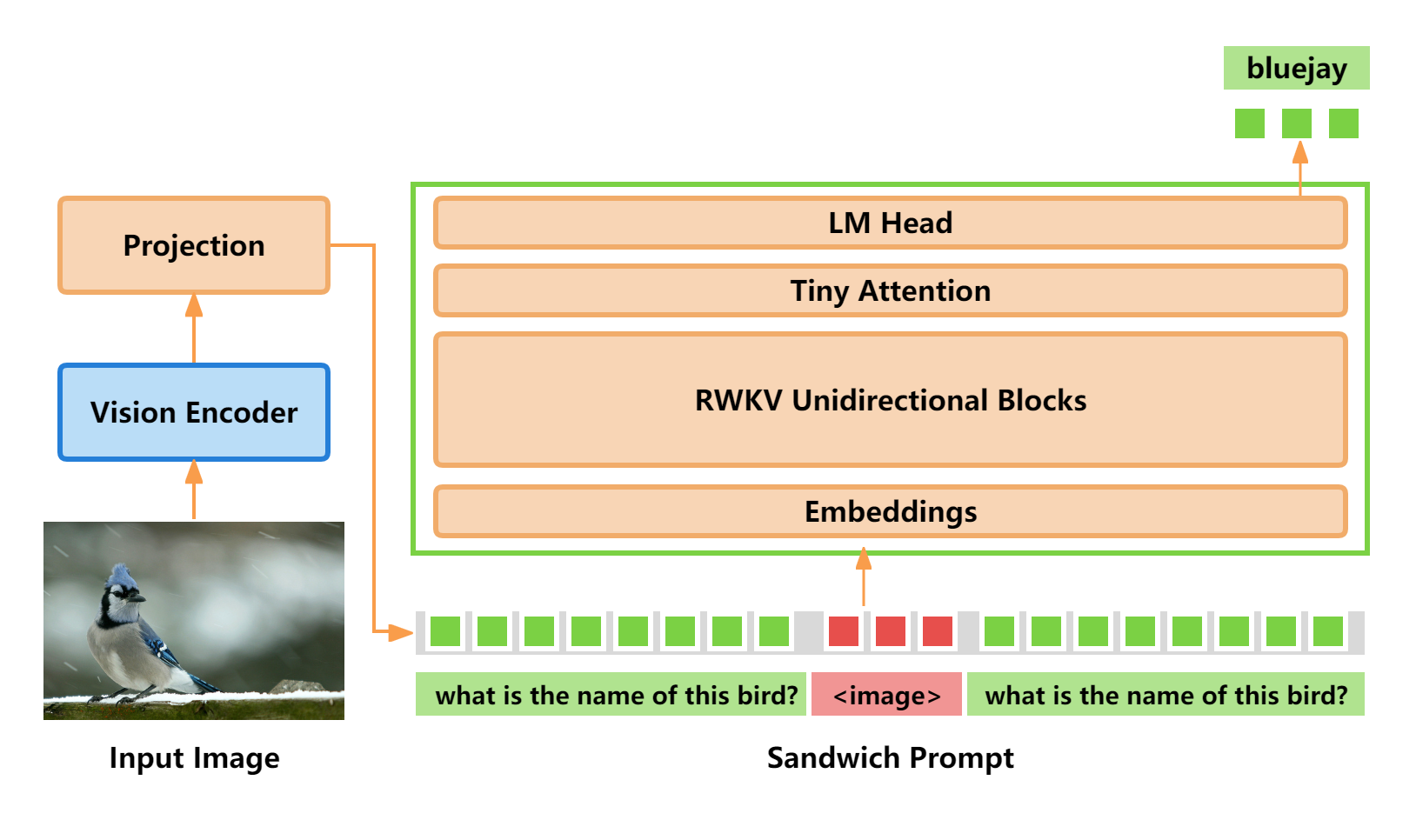}
    \vspace{-15pt}
    \caption{VisualRWKV Hybrid: Add a Tiny Attention Layer on the top of RWKV Blocks.}
    \label{fig:tiny-attention}
\end{figure*}

We have preliminarily explored the feasibility of the VisualRWKV hybrid model. The hybrid model refers to the combined use of RWKV and Attention. As shown in the Figure \ref{fig:tiny-attention}, we have simply added a layer of Tiny Attention on the top of the RWKV blocks. The parameter count of Tiny Attention is smaller than that of the standard Attention, and it does not include an FFN layer.

The results of the VisualRWKV hybrid are presented in Table \ref{tab:tiny_attention}. It can be observed that there is an improvement over the baseline model without tiny attention.
Considering the minimal increase in the number of parameters, this improvement is quite significant.
Additionally, we found that the hybrid model equipped with tiny attention is more robust to the number of image tokens.
These results suggest the incorporation of more Attention modules in future work may lead to further enhancements and enable the construction of superior Hybrid models.

\begin{table}[htb]\centering
\resizebox{0.65\textwidth}{!}{%
\begin{tabular}{lcccccc}\toprule
Method & Size &Image Tokens&ScienceQA&TextVQA &GQA \\\midrule
\multirow{8}{*}{VisualRWKV-Base} & \multirow{8}{*}{7B} & 577 & 65.2 & 50.25 & 50.5 \\
 &  & 145 & 64.90 & 46.38 & 47.47 \\
 &  & 65 & 64.40 & 44.58 & 45.09 \\
 &  & 37 & 64.11 & 44.01 & 44.78 \\
 &  & 17 & 63.86 & 43.61 & 44.57 \\
 &  & 10 & 63.26 & 43.27 & 44.37 \\
 &  & 5 & 62.87 & 43.03 & 44.08 \\
 &  & 1 & 60.34 & 41.72 & 36.09 \\
 \midrule
\multirow{8}{*}{VisualRWKV-Hybrid} & \multirow{8}{*}{7B} &577&\textbf{67.38} 
&\textbf{50.97} & 49.96 \\
 &  & 145 & 66.83 & 47.13 & 46.20 \\
 &  & 65 & 65.44 & 45.63 & 45.03 \\
 &  & 37 & 65.39 & 45.47 & 44.81 \\
 &  & 17 & 64.40 & 45.07 & 44.65 \\
 &  & 10 & 64.06 & 44.79 & 44.44 \\
 &  & 5 & 63.26 & 44.75 & 43.98 \\
 &  & 1 & 63.11 & 44.71 & 43.76 \\
\bottomrule
\end{tabular}
}
\caption{Results of VisualRWKV Hybrid model on 3 benchmarks. The prompting method used here is the sandwich prompt.}
\label{tab:tiny_attention}
\end{table}

\section{Use of AI Assistants}
In this research, an AI writing assistant is solely employed for the purposes of paraphrasing, spell-checking, and enhancing the author's original content, and it does not introduce any novel content.

\end{document}